\documentclass[10pt,twocolumn,letterpaper]{article}

\usepackage[pagenumbers]{cvpr} 

\usepackage{multirow}

\definecolor{cvprblue}{rgb}{0.21,0.49,0.74}
\usepackage[pagebackref,breaklinks,colorlinks,allcolors=cvprblue]{hyperref}

\usepackage{color}
\usepackage{pifont}
\usepackage{xcolor}
\usepackage{colortbl}

\usepackage{arydshln}

\title{Affordance Field Intervention: Enabling VLAs to Escape Memory Traps in Robotic Manipulation}

\author{
Siyu Xu$^{1}$ \quad Zijian Wang$^{1}$ \quad Yunke Wang$^{1}$ \quad Chenghao Xia$^{1}$ \quad Tao Huang$^{2}$ \quad Chang Xu$^{1}$ \\
$^1$School of Computer Science, The University of Sydney \\ $^2$John Hopcropt Center for Computer Science, Shanghai Jiao Tong University \\
\texttt{\small \{s.xu,yunke.wang,c.xu\}@sydney.edu.au}
}

\begin{document}

\twocolumn[{ 
  \maketitle  
  \begin{center}  
 \vskip -0.1in
    \includegraphics[width=1.0\linewidth]{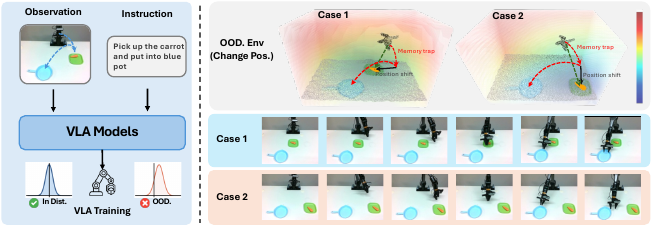}
    \captionof{figure}{\textbf{Memory Trap in VLAs.} VLAs often fail under distribution shifts. In such a case, they replay trajectories memorized during training instead of adapting to the updated scene. In both Case 1 and Case 2, even though the target object moves, the VLA drives the end-effector toward the original location, ignoring new spatial cues.}
    \label{figs:start}
  \end{center}
}] 

\begin{abstract}
Vision-Language-Action (VLA) models have shown great performance in robotic manipulation by mapping visual observations and language instructions directly to actions. 
However, they remain brittle under distribution shifts: when test scenarios change, VLAs often reproduce memorized trajectories instead of adapting to the updated scene, which is a failure mode we refer to as the ``Memory Trap''. 
This limitation stems from the end-to-end design, which lacks explicit 3D spatial reasoning and prevents reliable identification of actionable regions in unfamiliar environments. 
To compensate for this missing spatial understanding, 3D Spatial Affordance Fields (SAFs) can provide a geometric representation that highlights where interactions are physically feasible, offering explicit cues about regions the robot should approach or avoid. 
We therefore introduce Affordance Field Intervention (AFI), a lightweight hybrid framework that uses SAFs as an on-demand plug-in to guide VLA behavior. Our system detects memory traps through proprioception, repositions the robot to recent high-affordance regions, and proposes affordance-driven waypoints that anchor VLA-generated actions. 
A SAF-based scorer then selects trajectories with the highest cumulative affordance. Extensive experiments demonstrate that our method achieves an average improvement of 23.5\% across different VLA backbones ($\pi_{0}$ and $\pi_{0.5}$) under out-of-distribution scenarios on real-world robotic platforms, and 20.2\% on the LIBERO-Pro benchmark, validating its effectiveness in enhancing VLA robustness to distribution shifts.
\end{abstract}    
\section{Introduction}
\label{sec:intro}
Vision-Language-Action (VLA) models~\cite{kim2024openvla, brohan2024rt, firoozi2025foundation} have emerged as powerful motion planners in robotic manipulation, directly mapping visual observations and linguistic instructions to executable action sequences and enabling seamless interaction with objects in diverse environments. By leveraging large-scale pre-trained vision-language foundations, these end-to-end neural architectures facilitate generalization across tasks, such as grasping and object rearrangement, without requiring extensive task-specific engineering~\cite{black2024pi_0, intelligence2025pi_, khazatsky2024droid}. 

Despite these advancements, VLA models encounter substantial challenges in generalization, frequently succumbing to a ``Memory Trap''~\cite{zhou2025libero, fei2025libero, fang2025intention}. In out-of-distribution (OOD) scenarios, such as significant perturbations in object positions, these models tend to rigidly reproduce trajectories memorized from the training data. Consequently, this suboptimal behavior leads to task failure, as the model directs the end-effector toward obsolete positions rather than the actual updated target. This limitation stems from VLA's end-to-end design, which implicitly fits mappings from vision-language inputs to actions based on training distributions, without explicitly perceiving and reasoning over 3D regions for interaction~\cite{sun2025geovla, zhen20243d, li2025pointvla}. As a result, VLAs lack the planning capabilities to generate actions that target the appropriate regions in unfamiliar environments. Instead, they rely on memorized trajectories from training, which fail to adapt to perturbations~\cite{liu2025can, grover2025enhancing}.

Recent works have introduced the concept of affordance to guide action planning. 
An object's affordance, described as ``opportunities of interaction'', provides direct and intuitive guidance in robot workspace~\cite{yang2023grounding, gibson1978ecological, ahn2022can, ju2024robo}. 
By leveraging multimodal understanding and reconstruction techniques, these methods generate 3D spatial affordance fields~(SAFs) that highlight actionable regions~\cite{huang2023voxposer, huang2024rekep}. 
Once the SAF identifies target endpoints, non-learning-based methods such as optimization-based trajectory planning can generate dynamically feasible action trajectories~\cite{ji2024graspsplats, simeonov2022neural, yue2025vapo}.
However, these VLM-based planning approaches~\cite{huang2023voxposer, huang2024rekep} suffer from low success rates in practice due to two critical limitations: (1) unreliable VLM-generated motion plans that lack fine-grained geometric understanding and often produce infeasible actions~\cite{rusu2010fast, bellemare2020autonomous}, (2) heavy reliance on task-specific prompt engineering to generate diverse constraints, which are brittle and lack transferability across different manipulation scenarios~\cite{toussaint2015logic, li2025ddn}.

To overcome these challenges, we propose Affordance Field Intervention (AFI), a novel hybrid framework that treats a 3D SAF as a plug-in for VLA-based action generation. The core of our method lies in how to leverage the SAF to help the VLA escape the memory trap and navigate toward high-affordance regions, thereby improving task success rates. First, we design a memory trap detection mechanism using robot proprioception. By monitoring end-effector motion patterns and goal progress, our system identifies when the VLA falls into rigid, memorized trajectories that fail to adapt to environmental changes. Second, upon detecting the memory trap, the AFI rolls back the EEF to a recent high-affordance position for safe repositioning under the guidance of the SAF. From there, it proposes intermediate waypoints as nearby high-affordance points that progressively guide toward the target region (e.g., updated object location). These waypoints act as geometric anchors, breaking the VLA's rigid memorization with explicit spatial cues.

Finally, the VLA is queried to generate action proposals conditioned on these SAF-guided waypoints.To ensure spatial optimality, the SAF acts as a scorer, evaluating and re-ranking the VLA's action candidates based on their projected trajectories' cumulative affordance values. The action with the highest affordance value, indicating alignment with favorable paths, is selected for execution. This closed-loop modular integration allows the VLA to leverage its semantic understanding and efficiency while being softly constrained by 3D geometry through grounded interventions. It enables adaptive navigation to high-affordance regions \textit{without parameter updates}, effectively bridging data-driven policies with interpretable planning.

We conduct extensive experiments on both real-world robotic platforms and simulation benchmarks to validate our proposed AFI. On real-world manipulation tasks using an AgileX Piper manipulator, our method achieves consistent improvements across four diverse tasks, with average success rate gains ranging from 17.0\% to 26.0\% over baseline VLA policies ($\pi_{0}$ and $\pi_{0.5}$) under various OOD scenarios including position shifts, color changes, object variations, and background shifts. Our framework also demonstrates model-agnostic generalization, with ensemble integration of multiple VLA backbones achieving up to 89.0\% success rate. On LIBERO-Pro~\cite{zhou2025libero} simulation with spatial perturbations, our method improves $\pi_{0.5}$ by 21.7\% (Spatial) and 16.8\% (Object). These results validate that AFI effectively mitigates the memory trap problem and enhances VLA robustness to distribution shifts without requiring model retraining or additional demonstration data.

\section{Related Work}
\label{sec:related_work}

\subsection{Vision-Language-Action Models}
\label{sec:vla_vulnerabilities}

Vision-Language-Action (VLA) models~\cite{kim2024openvla, brohan2024rt, firoozi2025foundation} have emerged as a promising approach for general-purpose robotic manipulation. 
These models frame robot control as a sequence modeling task, trained end-to-end via imitation learning on large datasets of camera observations, language instructions, and actions.
However, a significant body of recent literature highlights that these models suffer from poor generalization and inherent fragility when faced with spatial perturbations or novel environments~\cite{fang2025intention,fei2025libero,zhou2025libero}.
Research suggests that existing VLA models often struggle with robust action planning in unseen contexts, relying heavily instead on memorized or imitated trajectories from their training data distribution~\cite{Foster2024IsBC, Belkhale2023DataQI}.

Recent studies also explored the integration of reinforcement learning (RL) to enhance the generalization capabilities of VLA models~\cite{liu2025can, li2025simplevla, lu2025vla, ye2025vla}. By incorporating RL-based fine-tuning, these approaches aim to adapt VLA policies to diverse and unseen scenarios beyond mere imitation. However, a key challenge lies in obtaining reliable reward signals for RL training~\cite{li2025vla, jiang2025irl}, which often requires extensive human annotation or complex simulation environments. Moreover, scaling RL to large-scale datasets and diverse environments remains a significant challenge, limiting its practicality for real-world robotic applications~\cite{lu2025vla, zang2025rlinf}.

\subsection{Grounding 3D Affordance and Action Planning}
\label{sec:grounding_3d_affordance_and_action_planning}

Grounding 3D affordances plays a pivotal role in robotic manipulation by bridging language instructions with actionable spatial representations, enabling agents to infer object interactions (e.g., graspable regions or avoidance zones) directly in the 3D perceptual space~\cite{jiang2022ditto, ahn2022can}. 
Grounding 3D affordances explicitly encodes target object locations and interaction cues into dense spatial maps, allowing action planning to be offloaded to these affordance maps.

Building on this, VoxPoser~\cite{huang2023voxposer} is a novel framework that leverages large language models (LLMs) and vision-language models (VLMs) to compose composable 3D value maps for zero-shot manipulation tasks. 
Geomanip~\cite{tang2025geomanip} uses geometric constraints as general interfaces for robot manipulation, allowing agents to reason about object interactions and plan actions in a more principled manner. GIGA~\cite{jiang2021synergies} is a structured implicit representation that couples 3D reconstruction with an affordance field for 6‑DoF grasping, grounding action cues directly in local geometry. By training the shared implicit functions on self‑supervised trials, it improves occlusion‑robust grasp detection and allocates representation capacity toward graspable regions, enabling more reliable action planning in clutter.

\begin{figure}[t]
\centering
\includegraphics[width=0.95\linewidth]{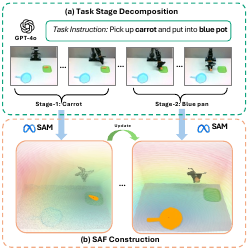}
   \caption{\textbf{Spatial Affordance Field (SAF) Construction.} (a) GPT-4o decomposes the task instruction into sequential stages and identifies the current target object (e.g., ``carrot'' or ``blue pan''). (b) The target text is fed to Grounded-SAM for segmentation, and the resulting 2D mask is back-projected into 3D space to construct the SAF, where color gradients indicate affordance values.}
\label{fig: preliminary}
\vspace{-5mm}
\end{figure}

\section{Preliminary}
\label{sec:preliminary}

\begin{figure*}[t]
\centering
\includegraphics[width=\linewidth]{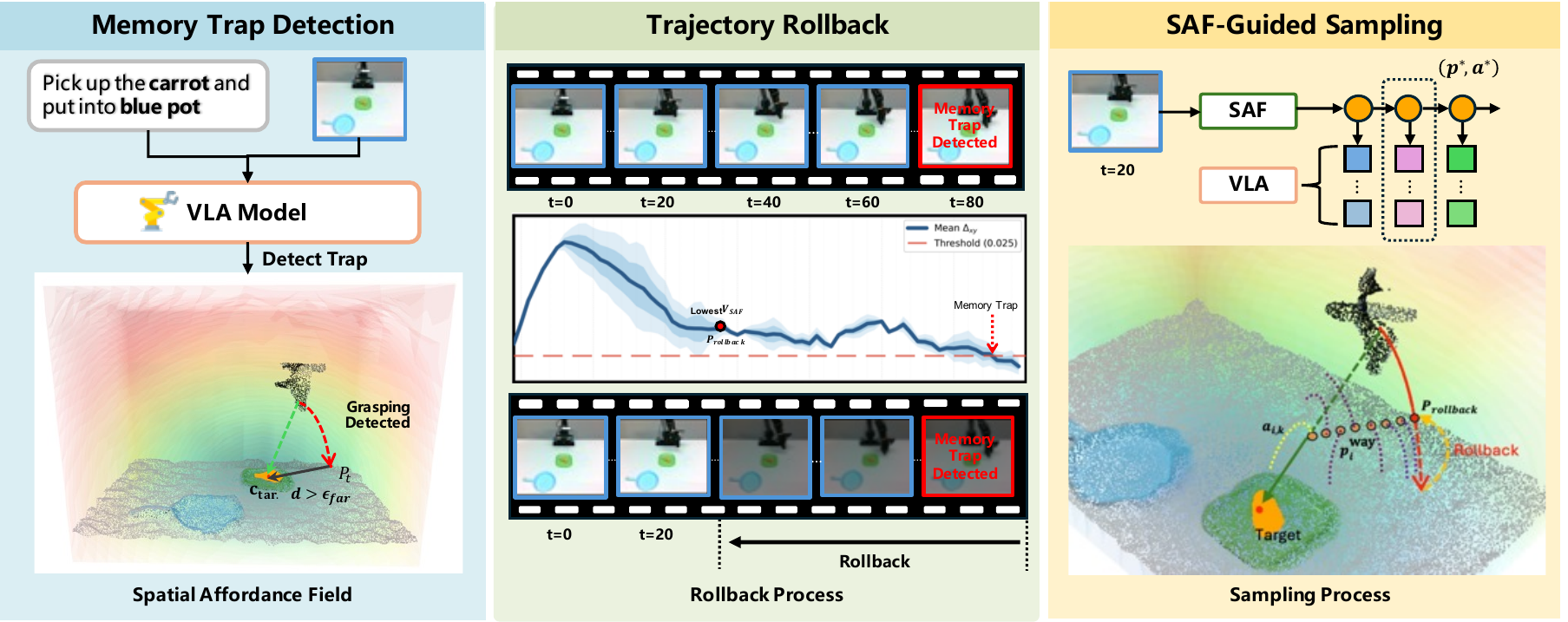}
\vspace{-3mm}
   \caption{\textbf{Overview of Affordance Field Intervention (AFI).} (1) \textit{Memory Trap Detection:} SAF evaluates VLA-predicted actions and detects memory traps by monitoring end-effector velocity and distance to target. (2) \textit{Trajectory Rollback:} Upon detection, the robot rolls back to the historical position with lowest SAF cost before grasping attempts. (3) \textit{SAF-Guided Sampling:} VLA generates trajectory candidates at SAF-sampled waypoints, and the trajectory with lowest cumulative SAF cost is selected for execution.}
\label{fig: overview}
\vspace{-5mm}

\end{figure*}

In this section, we introduce the background knowledge of VLA models and affordance field construction.

\subsection{VLA Models}
VLA models $\pi_{\text{VLA}}$ take a 2D image $I_t^{rgb}$ at timestep $t$ and language instructions $\tau$ as inputs to propose actions for environment interaction. 
These actions are executed by the controller, results in an end-effector~(EEF) displacement $\Delta \mathbf{d}_t$ in the 3D workspace, i.e.,
\begin{equation}
    a_t \sim \pi_{\text{VLA}}(I_t^{rgb}, \tau), \quad \Delta \mathbf{d}_t = \text{controller}(a_t),
\end{equation}
VLA models are typically trained using imitation learning, memorizing direct mappings from vision-language inputs to actions based on the training distribution. 
However, in out-of-distribution~(OOD) scenarios, the VLA model continues to propose actions that guide the end-effector along the memorized trajectories from the training set, rather than adapting to the perturbed environment.

\subsection{Affordance Field Construction}
In this subsection, we introduce the two-stage pipeline for constructing the 3D spatial affordance field~(SAF).
\subsubsection{Inferring Affordance via VLM}
\label{sec:affordance_inference}
To infer affordances from language instructions, we first identify task-relevant objects and then project them into the robot's workspace. 
Given a high-level task instruction $\tau$ (e.g., ``place the mug on the table'') and the current RGB observation $I_t^{rgb}$, we leverage a vision-language model (VLM) to parse the task into a sequence of semantic stages. 
Specifically, we use GPT-4o to decompose $\tau$ into temporally ordered sub-goals (e.g., \textit{pick} $\rightarrow$ \textit{move} $\rightarrow$ \textit{place}) and extract stage-wise target tokens (e.g., ``mug'', ``table''). 
These target tokens serve as text prompts for subsequent open-vocabulary object detection.

Then for each sub-goal, we apply Grounded-SAM \cite{ren2024grounded} to generate a 2D segmentation mask $M_{\text{target}} \in \{0, 1\}^{H \times W}$ corresponding to the target object. 
By combining $M_{\text{target}}$ with the depth map $I_t^{depth}$ and camera intrinsic parameters, we back-project the masked region into 3D space to obtain the target point cloud $\mathcal{P}_{\text{target}} = \{\mathbf{p}_i \in \mathbb{R}^3\}$. 
Notably, when the VLM detects a sub-goal transition during execution, the system automatically updates the target identification to reflect the new semantic focus, allowing the spatial affordance field to adapt dynamically throughout multi-stage manipulation tasks.

\subsubsection{Grounding Affordance in 3D Space}
\label{sec:affordance_field_construction}

We discretize the robot's workspace into an $N \times N \times N$ voxel grid $\mathcal{V}$, where each voxel $v_{ijk}$ represents a small cuboid region in 3D space. 
Both the scene point cloud $\mathcal{P}_{\text{scene}}$ (obtained from the full RGB-D observation) and the target object point clouds $\mathcal{P}_{\text{target}}$ are projected onto this grid.
Within this voxelized representation, we construct two complementary geometric subfields for each target object:

\textbf{1) Target Guidance Field} $V_{\text{target}}$: 
This field encodes spatial attraction toward the target. 
For each voxel $v_{ijk}$, we compute its Euclidean distance $d(v_{ijk}, \mathbf{c}_{\text{target}})$ to the target object centroid $\mathbf{c}_{\text{target}} = \frac{1}{|\mathcal{P}_{\text{target}}|}\sum_{\mathbf{p} \in \mathcal{P}_{\text{target}}} \mathbf{p}$. After applying distance transform, regions farther from the target receive higher values, which contribute to higher cost in the final affordance field, encouraging the EEF to approach the goal region.

\textbf{2) Obstacle Avoidance Field} $V_{\text{obst}}$: 
This field encodes repulsion from scene obstacles. Voxels occupied by $\mathcal{P}_{\text{scene}}$ or near obstacles are assigned high values to discourage collision. 
To prevent overly conservative behavior, we apply heuristic masking: (1) we exempt the immediate vicinity of the EEF, allowing close-range manipulation, and (2) we create a buffer zone around the target object, permitting approach actions necessary for grasping.

The final spatial affordance field $V_{\text{SAF}}$ is obtained by fusing these subfields via weighted linear combination:
\begin{equation}
    V_{\text{SAF}} = w_{\text{target}} V_{\text{target}} + w_{\text{obst}} V_{\text{obst}},
\end{equation}
where $w_{\text{target}}$ and $w_{\text{obst}}$ are hyperparameters balancing target attraction and obstacle avoidance. To ensure smooth spatial gradients suitable for trajectory evaluation, we apply Euclidean distance transform to both subfields, followed by Gaussian smoothing with kernel size $\sigma$. Finally, $V_{\text{SAF}}$ is normalized to $[0, 1]$, yielding a continuous affordance cost field where lower values indicate more favorable regions (near targets and away from obstacles) for action selection.

\section{Method}
\label{sec:method}

In this section, we introduce our proposed AFI that integrates the spatial affordance field (SAF) into the VLA workflow to help the VLA model escape the memory trap in out-of-distribution scenarios. 
We first describe how to detect whether the robot has fallen into the memory trap. Then, we explain how to integrate the SAF to intervene in the VLA workflow for escaping the memory trap.

\subsection{Memory Trap Detection}
We monitor the robot's execution status at each timestep to detect potential memory traps. 
A memory trap is triggered when two conditions are simultaneously met: (1) the end-effector displacement $\|\mathbf{p}_t - \mathbf{p}_{t-\Delta t}\|$ falls below a threshold $\epsilon_{\text{stuck}}$ over a time window $\Delta t$, and (2) the distance to the target $\|\mathbf{p}_t - \mathbf{c}_{\text{target}}\|$ exceeds a threshold $\epsilon_{\text{far}}$.
The first condition detects when the robot enters a quasi-static state, which typically indicates either fine-grained manipulation near the target (e.g., grasping) or getting stuck in an undesired configuration. 
The second condition disambiguates these two scenarios: if the end-effector becomes stationary while still far from the target, it suggests the robot is stuck or performing incorrect actions (e.g., grasping the wrong object), rather than executing the intended fine manipulation. 
This dual-criterion detection ensures we intervene only when genuine memory traps occur, avoiding false positives during legitimate stationary behaviors near the goal.

\subsection{Affordance Field Intervention}
\label{sec:saf_intervention_mechanism}
Upon detecting a memory trap, the spatial affordance field (SAF) immediately comes into play through a targeted intervention mechanism, guiding the escape from the trap and steering the VLA model away from its memorized trajectories. 
This intervention begins with a guided rollback to a safer historical position, followed by a tree-based, SAF-guided search that samples intermediate waypoints toward high-affordance regions and integrates VLA-generated trajectories for task-directed refinement.

\subsubsection{Historical Rollback via Affordance}

We maintain a history buffer $N$-step long $\mathbf{P}_{\text{hist}} = \{\mathbf{p}_{t-n}, \ldots, \mathbf{p}_{t-1}\}$ of recent end-effector positions.
The rollback target is selected as the historical point with the lowest affordance cost:
\begin{equation}
\mathbf{p}_{\text{rollback}} = \arg\min_{\mathbf{p} \in \mathbf{P}_{\text{hist}}} V_{\text{SAF}}(\mathbf{p}),
\end{equation}
where $V_{\text{SAF}}(\mathbf{p})$ queries the SAF at position $\mathbf{p}$, with lower values indicating lower cost (safer and more task-relevant regions).
The robot executes a short rollback trajectory to return to this configuration $\mathbf{p}_{\text{rollback}}$, which serves as the root node for subsequent tree-based trajectory extension. 
This step repositions the robot to a safer, low-cost state, mitigating the immediate effects of the memory trap.

\subsubsection{Hierarchical Exploration for Optimal Trajectories}

Starting from the rollback position $\mathbf{p}_{\text{rollback}}$, we construct a two-stage tree-based exploration process to sample intermediate waypoints toward high-affordance regions and integrate VLA-generated trajectories for task-directed refinement.

\textbf{Stage 1: Local SAF-Guided Sampling of Intermediate Waypoints.}
We perform a local spatial search to identify $N$ promising intermediate waypoints in the vicinity. 
Specifically, we sample candidate positions from a local neighborhood $\mathcal{N}(\mathbf{p}_{\text{rollback}}, r)$ with radius $r$ and select those with the lowest cost values:
\begin{equation}
    \{\mathbf{p}_i^{\text{way}}\}_{i=1}^N = \underset{\mathbf{p} \in \mathcal{N}(\mathbf{p}_{\text{rollback}}, r)}{\arg\min^N} \; V_{\text{SAF}}(\mathbf{p}),
\end{equation}
where $\arg\min^N$ selects the $N$ positions with minimum cost values. 
These waypoints form the first-level child nodes in the trajectory extension tree, representing spatially favorable intermediate targets that prioritize low-cost regions near targets and away from obstacles.

\textbf{Stage 2: Trajectory Generation via VLA at Sampled Waypoints.}
The robot sequentially navigates to each waypoint $\mathbf{p}_i^{\text{way}}$ and queries the VLA policy $\pi_{\text{VLA}}$ to produce $K$ diverse action candidates $\{\mathbf{a}_{i,k}\}_{k=1}^K$ based on the updated observation $I_t^{\text{rgb}}$ and task instruction $\tau$. 
For stochastic policies (e.g., diffusion-based), candidates are sampled with different noise temperature or seeds.

Each action candidate $\mathbf{a}_{i,k}$ represents an action chunk consisting of a sequence of joint states over horizon $H$.
We apply forward kinematics to convert these joint states into the corresponding end-effector trajectory $\boldsymbol{\xi}_{i,k} = \{\mathbf{p}_j^{i,k}\}_{j=1}^{H}$, where $\mathbf{p}_j^{i,k}$ denotes the end-effector position at timestep $j$.
The cumulative affordance cost is:

\begin{equation}
    \mathcal{V}(\boldsymbol{\xi}_{i,k}) = \sum_{j=1}^{H} V_{\text{SAF}}(\mathbf{p}_j^{i,k}).
\end{equation}

This yields $N \times K$ evaluated candidates forming the leaf nodes of the trajectory tree. 
We then select the globally optimal trajectory by minimizing the cumulative cost:
\begin{equation}
    \boldsymbol{\xi}^* = \arg\min_{i,k} \mathcal{V}(\boldsymbol{\xi}_{i,k}).
\end{equation}
The robot executes $\boldsymbol{\xi}^*$ by navigating to its corresponding waypoint and following the associated VLA actions. This hierarchical, exploration-driven approach effectively combines the spatial reasoning of SAF (for waypoint selection) with the task-specific capabilities of VLA (for trajectory completion), enabling robust recovery from diverse failure modes while incorporating real-world perceptual feedback.

\section{Experiments}
\label{sec:experiments}

\begin{table*}[!t]
\centering
\caption{Real-world experimental results on AgileX Piper manipulator across four manipulation tasks. We report success rates over 20 trials for each scenario. Our AFI framework achieves consistent improvements across all distribution shifts and VLA backbones. See Appendix~\ref{sec:appendix_tasks} for detailed task descriptions.}
\label{tab:real_world_results}
\setlength{\tabcolsep}{2.8mm}{
\begin{tabular}{ll|c|cccc|c}
\hline
\textbf{Task} & \textbf{Method} & \textbf{In Dist.} & \textbf{Position} & \textbf{Color} & \textbf{Task} & \textbf{Background} & \textbf{Average SR.} \\
\hline
\multirow{3}{*}{\textbf{Place Carrot}} 
& ReKep~\cite{huang2024rekep} & 8/20 & 7/20 & 9/20 & 5/20 & 7/20 & 36.0\% \\
& $\pi_{0}$ & 17/20 & 6/20 & 13/20 & 15/20 & 10/20 & 61.0\% \\
& \cellcolor[gray]{0.9}$\pi_{0}$-AFI (Ours) & \cellcolor[gray]{0.9}\textbf{20/20} & \cellcolor[gray]{0.9}\textbf{13/20} & \cellcolor[gray]{0.9}\textbf{17/20} & \cellcolor[gray]{0.9}\textbf{18/20} & \cellcolor[gray]{0.9}\textbf{19/20} & \cellcolor[gray]{0.9}\textbf{87.0\%} ($\uparrow$26.0\%) \\
\hdashline
\multirow{2}{*}{\textbf{Remove Lid}} 
& $\pi_{0}$ & 20/20 & 8/20 & 17/20 & 5/20 & 13/20 & 63.0\% \\
& \cellcolor[gray]{0.9}$\pi_{0}$-AFI (Ours) & \cellcolor[gray]{0.9}\textbf{20/20} & \cellcolor[gray]{0.9}\textbf{12/20} & \cellcolor[gray]{0.9}\textbf{19/20} & \cellcolor[gray]{0.9}\textbf{11/20} & \cellcolor[gray]{0.9}\textbf{18/20} & \cellcolor[gray]{0.9}\textbf{80.0\%} ($\uparrow$17.0\%) \\
\hdashline
\multirow{2}{*}{\textbf{Slot Pen}}
& $\pi_{0}$ & 16/20 & 11/20 & 13/20 & 15/20 & 5/20 & 60.0\% \\
& \cellcolor[gray]{0.9}$\pi_{0}$-AFI (Ours) & \cellcolor[gray]{0.9}\textbf{19/20} & \cellcolor[gray]{0.9}\textbf{16/20} & \cellcolor[gray]{0.9}\textbf{16/20} & \cellcolor[gray]{0.9}\textbf{19/20} & \cellcolor[gray]{0.9}\textbf{12/20} & \cellcolor[gray]{0.9}\textbf{82.0\%} ($\uparrow$22.0\%) \\
\hdashline
\multirow{5}{*}{\textbf{Stack Tape}}
& $\pi_{0}$ & 18/20 & 9/20 & 16/20 & 13/20 & 8/20 & 64.0\% \\
& \cellcolor[gray]{0.9}$\pi_{0}$-AFI (Ours) & \cellcolor[gray]{0.9}\textbf{20/20} & \cellcolor[gray]{0.9}\textbf{15/20} & \cellcolor[gray]{0.9}\textbf{20/20} & \cellcolor[gray]{0.9}\textbf{17/20} & \cellcolor[gray]{0.9}\textbf{14/20} & \cellcolor[gray]{0.9}\textbf{86.0\%} ($\uparrow$22.0\%) \\
\cline{2-8}
& $\pi_{0.5}$ & 20/20 & 7/20 & 17/20 & 10/20 & 7/20 & 61.0\% \\
& \cellcolor[gray]{0.9}$\pi_{0.5}$-AFI (Ours) & \cellcolor[gray]{0.9}\textbf{20/20} & \cellcolor[gray]{0.9}\textbf{14/20} & \cellcolor[gray]{0.9}\textbf{19/20} & \cellcolor[gray]{0.9}\textbf{15/20} & \cellcolor[gray]{0.9}\textbf{14/20} & \cellcolor[gray]{0.9}\textbf{82.0\%} ($\uparrow$21.0\%) \\
& \cellcolor[gray]{0.9}$\pi_{0}$+$\pi_{0.5}$-AFI (Ours) & \cellcolor[gray]{0.9}\textbf{20/20} & \cellcolor[gray]{0.9}\textbf{16/20} & \cellcolor[gray]{0.9}\textbf{20/20} & \cellcolor[gray]{0.9}\textbf{16/20} & \cellcolor[gray]{0.9}\textbf{17/20} & \cellcolor[gray]{0.9}\textbf{89.0\%} ($\uparrow$25.0\%) \\
\hline
\end{tabular}
}
\vskip -0.15in
\end{table*}

\subsection{Experimental Settings}

\noindent\textbf{Environments.}
We evaluate our framework on both real-world robotic platforms and simulated environments. For real-world experiments, we employ an AgileX Piper manipulator equipped with two Intel RealSense D435 cameras. 
For simulation experiments, we utilize the LIBERO benchmark~\cite{liu2023libero} with spatial perturbations following LIBERO-Pro~\cite{zhou2025libero} to assess OOD generalization. Detailed hardware configuration, SAF construction procedures, and training details are provided in Appendix~\ref{sec:appendix_implementation}.

\noindent\textbf{Baselines.}
We compare our method against pre-trained VLA models without 3D SAF guidance: $\pi_{0}$~\cite{black2024pi_0} and $\pi_{0.5}$~\cite{intelligence2025pi_} in real-world experiments, and the officially released $\pi_{0.5}$-LIBERO checkpoint in simulation. We also compare with ReKep~\cite{huang2024rekep}, a training-free VLM-based planner, to demonstrate the advantages of our hybrid VLA+SAF approach.

\noindent\textbf{Tasks.}
We evaluate on  four real-world manipulation tasks covering diverse primitives: (1) \textit{Place Carrot}: picking up a carrot and placing it in a pot; (2) \textit{Remove Lid}: removing a lid from a pot and placing it on a platter; (3) \textit{Slot Pen}: inserting a pen into a holder; (4) \textit{Stack Tape}: stacking one tape roll on top of another. 
Each task is evaluated over 20 trials under five test conditions: in-distribution (ID) and four OOD scenarios involving position shifts ($\pm$5-15cm), color/appearance changes, object property variations, and background shifts. 
Task-specific descriptions, training data details, and OOD scenario definitions are provided in Appendix~\ref{sec:appendix_tasks} and Figure \ref{fig:app_exps}.

\begin{figure}[t]
\centering
\includegraphics[width=0.99\linewidth]{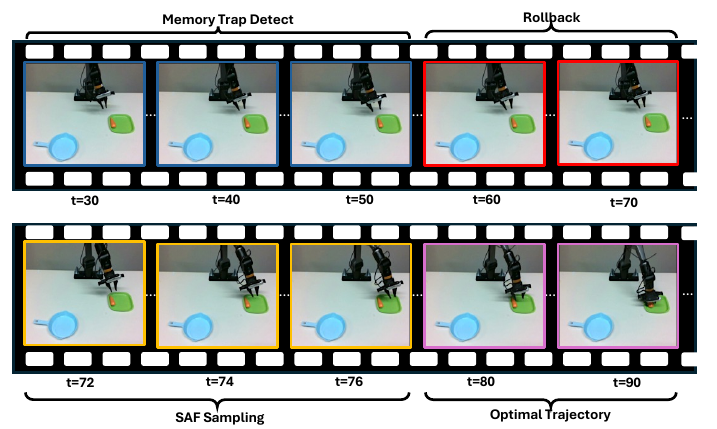}
\caption{\textbf{Real-world AFI execution rollout.} Top: Memory trap detected at $t=50$ when approaching wrong location, followed by rollback to low-cost historical position. Bottom: SAF-guided sampling ($t=70$-$79$) generates trajectory candidates; optimal trajectory (green) is selected and executed ($t=80$-$90$) for successful completion.}
\label{fig:real_saf}
\vskip -0.25in
\end{figure}

\subsection{Main Results in Real World}

Table~\ref{tab:real_world_results} presents comprehensive results across four manipulation tasks and various distribution shift scenarios. 
Our AFI framework achieves consistent improvements across all tasks and VLA backbones, with average success rate gains ranging from 17.0\% to 26.0\% over baseline policies.

\noindent\textbf{Consistent Improvements Across Tasks.}
Our method demonstrates robust performance gains across all manipulation primitives. 
For \textit{Place Carrot}, $\pi_{0}$-AFI achieves 87.0\% average success rate versus 61.0\% baseline, with particularly strong improvements in Position shift (65\% vs 30\%) and Background shift (95\% vs 50\%) scenarios. 
For \textit{Stack Tape}, we observe 86.0\% success rate with $\pi_{0}$-AFI compared to 64.0\% baseline, achieving perfect success (100\%) in Color shift scenarios. 
The \textit{Slot Pen} task shows 82.0\% success (22.0\% gain), while \textit{Remove Lid} achieves 80.0\% (17.0\% gain). 
These results validate that our SAF-guided intervention effectively handles diverse manipulation primitives, including \textit{picking, placing, insertion, lid removal, and stacking} operations. Figure~\ref{fig:real_saf} visualizes a typical AFI execution rollout, demonstrating the complete intervention process from memory trap detection to successful task completion.

\noindent\textbf{Robustness to Challenging Distribution Shifts.}
The Task shift scenarios present the most challenging OOD conditions, involving physical property variations and distractors. 
Our method shows substantial improvements: 30.0\% gain for \textit{Remove Lid} with distractor avoidance (from 25\% to 55\%), 20.0\% gain for \textit{Slot Pen} with thinner pen insertion (from 75\% to 95\%), and 20.0\% gain for \textit{Stack Tape} with different tape types (from 65\% to 85\%). 
Position shift scenarios also benefit significantly, with average improvements of 25.0\% across tasks, demonstrating that our SAF provides effective 3D spatial reasoning to locate displaced objects.

\noindent\textbf{Comparison with Zero-Shot VLM Planner.} 
ReKep~\cite{huang2024rekep}, a training-free VLM-based planner, achieves only 36.0\% average success rate on the \textit{Place Carrot} task. 
While VLMs excel at semantic understanding, they lack fine-grained motion planning capabilities required for precise manipulation. 
In contrast, our hybrid approach achieves 87.0\% by leveraging VLMs' zero-shot grounding abilities to construct SAFs as guidance while relying on VLA models for robust action generation, combining their complementary strengths.

\noindent\textbf{Generalization Across VLA Backbones.}
We evaluate both $\pi_{0}$ and $\pi_{0.5}$ on the \textit{Stack Tape} task. 
Both achieve similar baseline performance (64.0\% and 61.0\%), and our AFI improves them to comparable levels (86.0\% and 82.0\%). 
This consistency validates the model-agnostic nature of our approach, demonstrating that our framework generalizes across different VLA architectures without requiring architecture-specific modifications.

\noindent\textbf{Ensemble Integration for Enhanced Performance.}
We further explore ensemble integration by combining action proposals from both $\pi_{0}$ and $\pi_{0.5}$ backbones, where our SAF-based scorer selects the optimal trajectory from the combined candidate pool. 
As shown in Table~\ref{tab:real_world_results}, the ensemble approach ($\pi_{0}$+$\pi_{0.5}$-AFI) achieves 89.0\% average success rate on \textit{Stack Tape}, outperforming both individual backbones ($\pi_{0}$-AFI: 86.0\%, $\pi_{0.5}$-AFI: 82.0\%) and representing a 25.0\% improvement over baseline $\pi_{0}$ (64.0\%). 
This demonstrates a key advantage of our framework: its model-agnostic design naturally supports multi-policy integration, leveraging complementary strengths from different VLA architectures without complex fusion mechanisms.

\begin{figure}[t]
\centering
\includegraphics[width=0.95\linewidth]{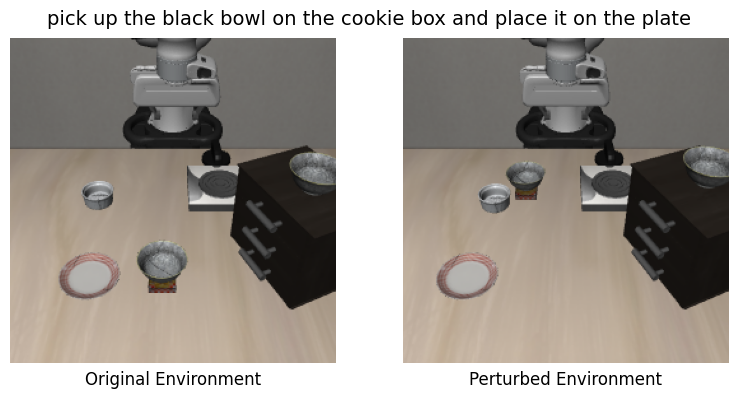}
\caption{Visualization of object position perturbations in LIBERO simulation, where the target object ``black bowl on the cookie box'' is displaced to significantly deviated positions.}
\vspace{-5mm}
\label{fig:libero_pro}
\end{figure}

\subsection{Main Results in Simulation}
For LIBERO simulation experiments, we evaluate our method on the LIBERO-Pro~\cite{zhou2025libero} benchmark by introducing spatial perturbations to target object positions in each subtask, following the OOD evaluation protocol from LIBERO-Pro. 
These perturbations are carefully designed to avoid rendering errors, prevent object collisions, and ensure no semantic contradictions with the task instructions~\cite{zhou2025libero}.

\noindent Table~\ref{tab:libero_pro_results} reports the success rates of our method compared to the baseline $\pi_{0.5}$ model on the LIBERO-Spatial and LIBERO-Object suites, broken down by each subtask. 
Our SAF-guided framework demonstrates substantial improvements across most subtasks, particularly in scenarios involving spatial perturbations, achieving an average success rate of 78.2\% on LIBERO-Spatial (vs. 52.4\% for $\pi_{0.5}$) and 82.5\% on LIBERO-Object (vs. 67.3\% for $\pi_{0.5}$). These results underscore the effectiveness of affordance-guided interventions in enhancing VLA generalization to out-of-distribution object configurations and positional shifts in simulation environments.

\begin{table}[!t]
    \centering
    \caption{Success rates of $\pi_{0.5}$-AFI and $\pi_{0.5}$ on the LIBERO-Spatial and LIBERO-Object, with object position perturbations introduced to subtask following the LIBERO-Pro protocol.}
    \label{tab:libero_pro_results}
  \setlength{\tabcolsep}{1.8mm}{
        \begin{tabular}{lcc}
            \hline
            \textbf{LIBERO-Spatial~(OOD)} & & \\
             \hline
            Task & $\pi_{0.5}$            & $\pi_{0.5}$-AFI \\
            \hdashline
            Pick(between(plate, ramekin), plate)        & 70.0\% & \textbf{82.0}\% \\
            Pick(next\_to(ramekin), plate)              & 22.0\% & \textbf{54.0}\% \\
            Pick(table\_center, plate)                  & 96.0\% & \textbf{98.0}\% \\
            Pick(on(cookie\_box), plate)                & 74.0\% & \textbf{88.0}\% \\
            Pick(on(ramekin), plate)                    & 26.0\% & \textbf{54.0}\% \\
            Pick(next\_to(cookie\_box), plate)          & 36.0\% & \textbf{72.0}\% \\
            Pick(next\_to(plate), plate)                & 54.0\% & \textbf{82.0}\% \\
            \hdashline
            Average                                     & 54.0\% & \textbf{75.7}\% \\
            \hline
            \textbf{LIBERO-Object~(OOD)} & & \\
            \hline
            Task & $\pi_{0.5}$ & $\pi_{0.5}$-AFI \\
            \hdashline
            Place(alphabet\_soup, basket)               & 42.0\% & \textbf{64.0}\% \\
            Place(bbq\_sauce, basket)                  & 54.0\% & \textbf{72.0}\% \\
            Place(butter, basket)                     & 78.0\% & \textbf{82.0}\% \\
            Place(chocolate\_pudding, basket)           & 88.0\% & \textbf{90.0}\% \\
            Place(cream\_cheese, basket)            & 42.0\% & \textbf{56.0}\% \\
            Place(ketchup, basket)                 & 46.0\% & \textbf{66.0}\% \\
            Place(milk, basket)                      & 88.0\% & \textbf{92.0}\% \\
            Place(orange\_juice, basket)            & 70.0\% & \textbf{80.0}\% \\
            Place(salad\_dressing, basket)           & 16.0\% & \textbf{64.0}\% \\
            Place(tomato\_sauce, basket)          & 40.0\% & \textbf{66.0}\% \\
            \hdashline
            Average                           & 56.4\% & \textbf{73.2}\% \\
            \bottomrule
        \end{tabular}
    }
    \vskip -0.2in
\end{table}

\noindent\textbf{Computational Efficiency.}
Our framework introduces minimal overhead to the baseline VLA inference. SAF reconstruction takes 120 ms per frame using Grounded-SAM and point cloud processing on an NVIDIA RTX 4090, while waypoint generation and action re-ranking add 15 ms total. This results in an end-to-end latency of 185 ms, suitable for 5 Hz control rates in real-world deployment. In contrast, pure optimization-based methods like MPC require 500+ ms per planning step, limiting their applicability to high-frequency manipulation.

\subsection{Ablations}

\begin{figure}[t]
\centering
\includegraphics[width=0.97\linewidth]{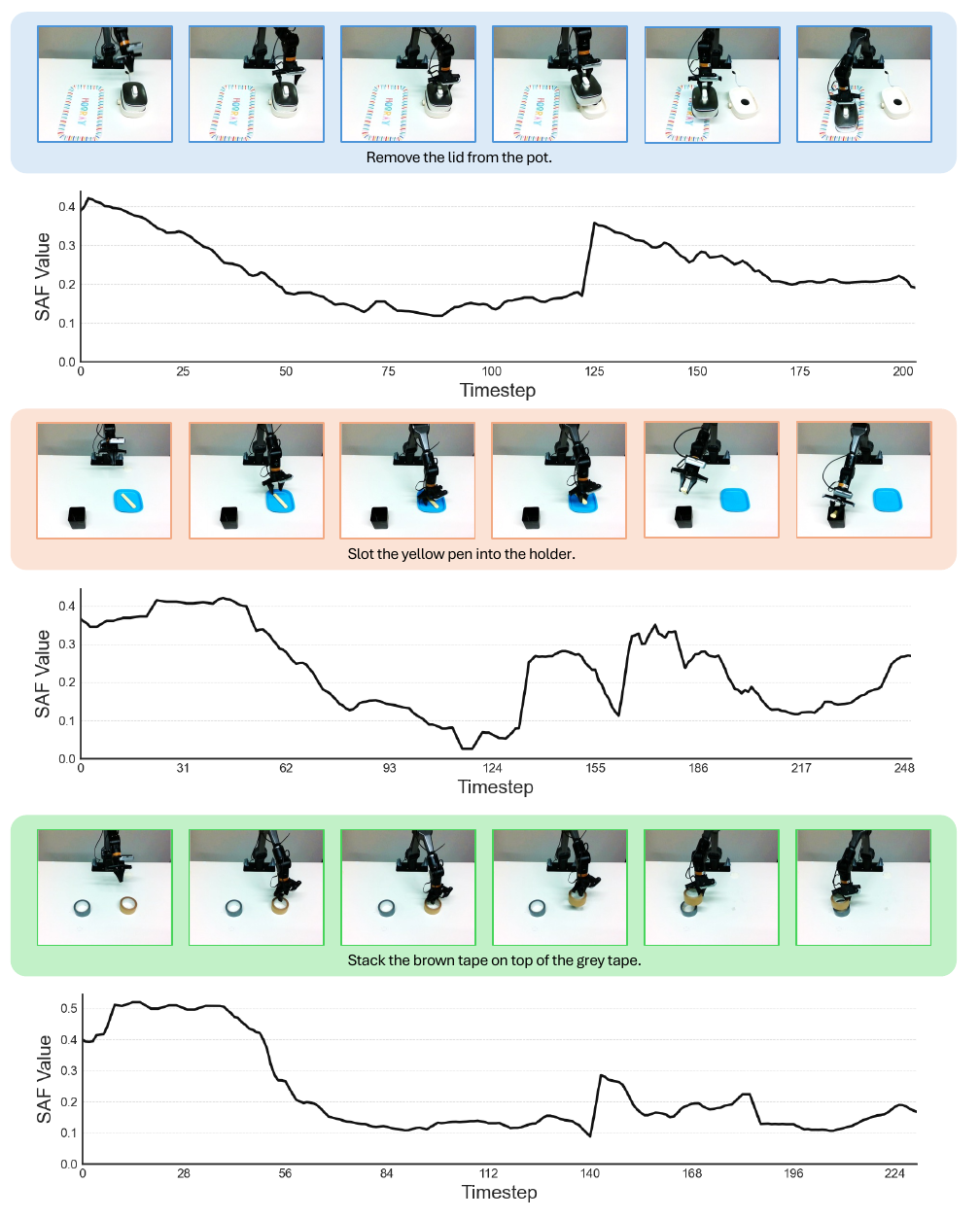}
\caption{\textbf{SAF value evolution across manipulation tasks.} Lower cost values indicate higher affordance. Costs decrease when approaching targets and update dynamically during stage transitions (e.g., picking to placing).}
\vspace{-6mm}
\label{fig:saf_value_vis}
\end{figure}

\noindent\textbf{SAF adaptation throughout task execution.}
To validate that our SAF accurately reflects task-relevant spatial information, we analyze its cost evolution throughout complete trajectories (Figure~\ref{fig:saf_value_vis}). 
The curves demonstrate that SAF cost decreases as the end-effector approaches target objects, validating that our affordance field accurately reflects spatial proximity to manipulation goals.
Notably, the cost values exhibit dynamic updates when transitioning between manipulation stages (e.g., from picking the lid to placing it in the platter), showing that our system adaptively adjusts the affordance field based on current task semantics.
This stage-aware adaptation is crucial for multi-step manipulation tasks, ensuring that the spatial guidance remains relevant throughout the entire execution sequence.

\noindent\textbf{Robustness to position shifts along different axes.}
To investigate VLA models' robustness to spatial perturbations, we systematically shift the target object along X-axis and Y-axis independently and jointly. Table~\ref{tab:position_ablation} presents results across six position configurations. $\pi_{0}$'s performance degrades catastrophically under single-axis shifts: success rates drop to 15\% for $\Delta X=+10$ cm and 5\% for $\Delta X=+15$ cm, revealing that VLA models memorize specific spatial patterns and fail when deviating along unseen directions. Interestingly, diagonal displacement ($\Delta X=+10$ cm, $\Delta Y=+10$ cm) maintains 30\% success, likely because such trajectories align with the training distribution's spatial coverage. Our method consistently improves robustness across all configurations (40\% for single-axis, 65\% for diagonal), validating that explicit 3D spatial guidance helps escape memory traps. However, extreme OOD scenarios ($\Delta X/Y=+15$ cm) show diminishing returns, highlighting that our approach complements rather than replaces the VLA's learned priors.

\begin{table}[h]
\centering
\caption{Ablation study on position shifts. We evaluate $\pi_{0}$ and $\pi_{0}$-AFI under different spatial displacements along X and Y axes (in centimeters).}
\label{tab:position_ablation}
\resizebox{\columnwidth}{!}{
\begin{tabular}{l|cccccc}
\hline
$(\Delta X, \Delta Y)$ & (0,0) & (+10,0) & (+15,0) & (0,+10) & (0,+15) & (+10,+10) \\
\hline
$\pi_{0}$ & 17/20 & 3/20 & 1/20 & 4/20 & 0/20 & 6/20 \\
$\pi_{0}$-AFI & \textbf{20/20} & \textbf{8/20} & \textbf{3/20} & \textbf{10/20} & \textbf{2/20} & \textbf{13/20} \\
\hline
\end{tabular}
}
\end{table}

\noindent\textbf{Ablation on key components.}
Table~\ref{tab:component_ablation} validates the importance of individual components on the position shift scenario. Removing rollback degrades performance from 65\% to 40\%, demonstrating its critical role. Without it, the VLA deviates too far from viable trajectories, and SAF-guided waypoints cannot recover from failed positions. Rollback repositions the end-effector to a recent high-affordance state, providing a safe starting point for effective redirection. Fixed-step interventions achieve at most 60\% (Step 30), underperforming our adaptive detection (65\%). This validates that real-time proprioceptive monitoring enables flexible intervention precisely when memory traps occur, optimizing both efficiency and success rate.

\begin{table}[t]
\small
\centering
\caption{Ablation study on key components: (1) rollback mechanism and (2) adaptive detection vs. fixed-step intervention. All experiments on position shift scenario over 20 trials.}
\label{tab:component_ablation}
  \setlength{\tabcolsep}{6.8mm}{
\begin{tabular}{l|cc}
\hline
Method & Success & Failure \\
\hline
$\pi_{0}$ & 6/20 & 14/20 \\
\rowcolor[gray]{0.9}
$\pi_{0}$-AFI & \textbf{13/20} & 7/20 \\
\hdashline
w/o Rollback & 8/20 & 12/20 \\
Fixed-step at 30 & 12/20 & 8/20 \\
Fixed-step at 60 & 11/20 & 9/20 \\
Fixed-step at 90 & 9/20 & 11/20 \\
\hline
\end{tabular}
}
\vspace{-3mm}
\end{table}

\noindent\textbf{Impact of waypoint proposal count.}
Table~\ref{tab:waypoint_count} examines how the number of waypoint proposals affects performance. With only 3 proposals, success rate remains low at 35\%, barely improving over the baseline. Optimal performance is achieved at 10 proposals (65\%), balancing spatial exploration and computational efficiency. Further increasing to 13 proposals shows marginal degradation (60\%), possibly due to over-exploration introducing suboptimal waypoints. These results highlight the importance of balanced sampling for effective coverage of high-affordance regions.

\begin{table}[t]
\centering
\caption{Ablation on waypoint count with different numbers of SAF-sampled candidates.}
\label{tab:waypoint_count}
\begin{tabular}{l|cccc}
\hline
Num of Waypoints & 3 & 8 & 10 & 13\\
\hline
Success Rate & 35.0\% & 50.0\% & 65.0\% & 60.0\%\\
\hline
\end{tabular}
\vspace{-5mm}
\end{table}
\section{Conclusion}
\label{sec:conclusion}

We identified the ``Memory Trap'' problem where VLAs rigidly reproduce memorized trajectories under distribution shifts. To address this, we proposed Affordance Field Intervention (AFI), which augments VLA models with explicit 3D spatial reasoning via Spatial Affordance Fields. AFI operates through proprioceptive memory trap detection, guided rollback to high-affordance positions, and hierarchical trajectory exploration via SAF-guided waypoint sampling. By treating SAF as an on-demand plug-in without modifying VLA parameters, our method is model-agnostic and applicable to any pre-trained VLA backbone.
Extensive experiments demonstrate significant improvements: 23.5\% average gain on real-world OOD scenarios and 20.2\% on LIBERO-Pro benchmark. The training-free nature makes it practical for deployment without additional data or fine-tuning. Our work shows that combining VLA policies with interpretable 3D spatial affordance fields offers a promising path toward more robust and generalizable robotic manipulation systems. 
{
    \small
    \bibliographystyle{ieeenat_fullname}
    \bibliography{main}
}

\clearpage
\appendix
\maketitlesupplementary

\section{Implementation Details}
\label{sec:appendix_implementation}

\noindent\textbf{Hardware Configuration.}
Our real-world experiment setup employs an \textit{AgileX Piper manipulator} equipped with two \textit{Intel RealSense D435 cameras}: one mounted on the wrist and another positioned in front of the robot. Both cameras are calibrated relative to the robot base frame to enable accurate 3D point cloud reconstruction. We utilize the calibrated front-mounted RealSense camera for scene reconstruction, operating at 30 Hz observation frequency. During data collection, we utilize RGB images from both viewpoints, while depth information is exclusively used at inference time to construct 3D point clouds for spatial affordance field generation.

\noindent\textbf{Spatial Affordance Field Construction.}
For affordance field construction, we apply the GPT-4o API to parse task instructions and identify manipulation stages. Open-vocabulary detection and target object tracking are performed locally on an NVIDIA GeForce GTX 1080Ti GPU using Grounded-SAM~\cite{ren2024grounded} to generate 2D instance segmentation masks. The complete SAF is published and updated as a ROS topic at 2 Hz frequency, ensuring real-time spatial reasoning without introducing latency to the original VLA inference pipeline.

\noindent\textbf{Control and Kinematics.}
Additionally, we deploy Curobo on the same 1080Ti GPU for forward and inverse kinematics computation. This enables bidirectional transformation: converting VLA-predicted joint states to end-effector spatial coordinates, and mapping SAF-sampled waypoints back to joint configurations. The kinematics computation introduces approximately 5ms latency and operates as a ROS service at 10Hz frequency. Overall, our SAF updates at 2Hz without adding overhead to the VLA policy inference pipeline, maintaining efficient real-time control.

\noindent\textbf{Training Details.}
For data collection, we use a master-follower teleoperation setup with an auxiliary AgileX Piper arm. The baseline VLA models ($\pi_{0}$ and $\pi_{0.5}$) are fully fine-tuned on collected demonstration trajectories for 30,000 steps with batch size 32 on a single NVIDIA H100 GPU. During inference, we sample 8 action chunks per query to ensure trajectory diversity. For stochastic action generation via flow matching, we set the initial noise sampling temperature (standard deviation) to 1.5 to encourage diverse proposals, which are then re-ranked by our SAF-based scorer.

\section{Real-world Task Settings}
\label{sec:appendix_tasks}

\begin{figure*}[!t]
\centering
\includegraphics[width=\linewidth]{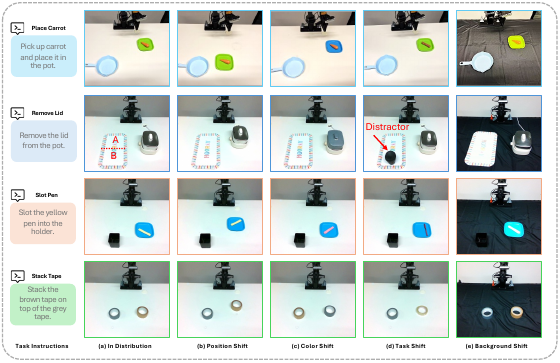}
\caption{\textbf{Visual illustration of OOD test scenarios.} Four manipulation tasks across five test conditions: (a) in-distribution setting, (b) position shift (objects displaced by $\pm$5-15cm), (c) color shift (object appearance change), (d) task shift (physical property variations or distractors), and (e) background shift (table surface color change from white to black). Each row shows a different task: Place Carrot (top), Remove Lid (second), Slot Pen (third), and Stack Tape (bottom).}
\label{fig:app_exps}
\end{figure*}

We evaluate our framework on four real-world manipulation tasks with varying complexity, each designed to test different manipulation primitives and robustness to distribution shifts. Each task is evaluated over 20 trials under five test conditions: in-distribution (ID) and four OOD scenarios (illustrated in Figure~\ref{fig:app_exps}).

\subsection{Task Description and Collection}

\noindent\textbf{Task 1 - Place Carrot.} The instruction is ``Pick up the carrot and place it in the pot.'' 
The objective is to pick up a carrot from a plate and place it into a pot. 
Success is defined as successfully placing the carrot into the pot. 
We collect 68 expert demonstration trajectories using a master-follower teleoperation setup.

\noindent\textbf{Task 2 - Remove Lid.} The instruction is ``Remove the lid from the pot.'' 
The objective is to remove the stainless steel lid from a pot and place it onto a nearby platter. 
Success is defined as placing the lid completely within the platter boundaries. 
We collect 78 expert demonstration trajectories with the pot position fixed. 
The platter is divided into two regions (A and B), and the lid placement is distributed across both regions, with 39 trajectories collected for each region.

\noindent\textbf{Task 3 - Slot Pen.} The instruction is ``Slot the yellow pen into the holder.'' 
The task requires inserting a yellow marker pen into a pen holder. 
Success is defined as the pen being fully inserted into the holder. 
We collect 77 expert demonstration trajectories for this task.

\noindent\textbf{Task 4 - Stack Tape.} The instruction is ``Stack the brown tape on top of the grey tape.'' 
This task involves placing a roll of brown tape on top of a roll of grey tape. 
Success is defined as the brown tape resting stably on the grey tape without falling. 
We collect 80 expert demonstration trajectories for this task.

\subsection{Out-of-Distribution Test Scenarios}

Following the evaluation protocol in our main experiments, we design four types of distribution shifts for each task. Figure~\ref{fig:app_exps} provides a visual illustration of these OOD test scenarios across all four tasks:

\noindent\textbf{Position Shift.} Objects are displaced within a 5-15cm radius from their training positions. 
For Place Carrot, the carrot is displaced by approximately 15 cm; 
for Remove Lid, we move the pot within 5cm; 
for Slot Pen, we move the blue plate containing the pen within 5cm; 
for Stack Tape, we move the tape placement area within 5cm.
    
\noindent\textbf{Color Shift.} For Place Carrot, the plate holding the carrot is changed from green to blue. 
For Remove Lid, the stainless steel lid is replaced with a grey-colored lid. 
For Slot Pen, the yellow marker is replaced with a pink one. 
For Stack Tape, the brown tape is replaced with a grey-colored tape.
    
\noindent\textbf{Task Shift.} For Place Carrot, the target object is replaced from carrot to sausage while maintaining the same task structure. 
For Remove Lid, we add a black cup as a distractor in the platter region and extend the instruction with the phrase ``Avoid the black cup.'' 
For Slot Pen, the yellow marker is replaced with a thinner red pen, requiring different insertion dynamics. 
For Stack Tape, the brown tape is replaced with a different type of tape (different shape and texture).
    
\noindent\textbf{Background Shift.} The table surface color is changed from white to black across all tasks.

\end{document}